\RequirePackage{etex} 
\documentclass[runningheads,twocolumn,a4paper,10pt]{llncs}

\usepackage{etex}
\reserveinserts{28}
\pdfoutput=1
\usepackage[a4paper, total={7in, 10in}]{geometry}
\usepackage{ textcomp }
\usepackage{ comment }
\usepackage[color=yellow!60,textsize=footnotesize,obeyDraft,draft]{todonotes}
\usepackage{colortbl}
\usepackage{booktabs}
\usepackage{capt-of}
\usepackage{makecell}
\usepackage{multirow}
\usepackage{listings}
\usepackage{graphicx}
\usepackage{amsmath}
\usepackage{caption}
\usepackage{subcaption}
\usepackage[font=footnotesize,labelfont=bf]{subcaption}
\usepackage[hidelinks]{hyperref}
\usepackage[capitalize]{cleveref}
\usepackage{tikz}
\usepackage{pgfplots}
\pgfplotsset{compat=newest}
\pgfplotsset{plot coordinates/math parser=false}
\usepackage[para,online,flushleft]{threeparttable}
\usepackage{wasysym}
\usepackage{amssymb}
\usetikzlibrary{arrows}
\usepackage{breakcites}
\usepackage{array}
\usepackage{color}
\usepackage{balance} 
\usepackage{lastpage}
\usepackage{dblfloatfix}
\usepackage{authblk}
\usepackage{siunitx}
\usepackage{xurl} 
\usepackage[shortlabels]{enumitem}
\usepackage{soul}
\usepackage{xcolor,listings}
\usepackage{relsize}
\usepackage{algorithm}
\usepackage{algpseudocode}

\usepackage{float}
\floatstyle{plaintop}
\restylefloat{table}
\usepackage{tikz}
\usetikzlibrary{shapes.geometric, arrows,calc}
\tikzstyle{startstop} = [rectangle, rounded corners, minimum width=3cm, minimum height=1cm,text centered, draw=black, fill=none]
\tikzstyle{io} = [trapezium, trapezium left angle=70, trapezium right angle=110, minimum width=3cm, minimum height=1cm, text centered, draw=black, fill=blue!30]
\tikzstyle{process} = [rectangle, minimum width=3.5cm, minimum height=1cm, text centered, draw=black, fill=orange!0]
\tikzstyle{neuron} = [circle, minimum size=3cm, text centered, draw=black, fill=orange!0]
\tikzstyle{decision} = [diamond, minimum width=3cm, minimum height=1cm, text centered, draw=black, fill=green!30]
\tikzstyle{arrow} = [thick,->,>=stealth]

\usepackage[maxnames=6,firstinits=true,doi=false,url=true,isbn=false]{biblatex}
\DeclareMathOperator*{\minimize}{min\text{ }}

\begin{document} 

\title{DIRA-SS: \textbf{D}ynamic Domain \textbf{I}ncremental \textbf{R}egularised \textbf{A}daptation - Self-Supervised}
\titlerunning{DIRA-SS}
\authorrunning{Ghobrial et al.}

\author{Abanoub Ghobrial, Kerstin Eder}

\institute{University of Bristol, Bristol, UK}
\tocauthor{Authors' Instructions}

\maketitle
\let\thefootnote\relax\footnotetext{
Abanoub Ghobrial (e-mail: abanoub.ghobrial@bristol.ac.uk)
and 
Kerstin Eder (e-mail: kerstin.eder@bristol.ac.uk) 
are with the Trustworthy Systems Lab, Department of Computer Science, University of Bristol, Merchant Ventures Building, Woodland Road, Bristol, BS8 1UB, United Kingdom. 
}

\makeatletter
\renewcommand\subsubsection{\@startsection{subsubsection}{3}{\z@}%
                       {-18\p@ \@plus -4\p@ \@minus -4\p@}%
                       {4\p@ \@plus 2\p@ \@minus 2\p@}%
                       {\normalfont\normalsize\bfseries\boldmath
                        \rightskip=\z@ \@plus 8em\pretolerance=10000 }}
\makeatother

\textbf{\textit{Abstract}--Autonomous systems (AS) often rely on Deep Neural Network (DNN) classifiers to operate in complex and dynamically changing environments. 
However, during operation, these classifiers may encounter domains that differ from those seen during development, causing performance degradation under distribution shift.
Removing systems from operation for labelled data collection and retraining is often impractical, particularly when adaptation must occur quickly and at scale. 
This paper introduces DIRA-SS, a self-supervised extension of Dynamic Incremental Regularised Adaptation (DIRA) that enables online domain adaptation using only a small number of unlabelled target-domain samples. 
DIRA-SS augments an existing classifier with an auxiliary retraining branch and adapts the shared feature representation through a rotation-prediction task, while elastic weight consolidation regularises important source-domain parameters to reduce destructive updates. 
This allows the model to benefit from transfer learning without requiring classification labels during operation. We evaluate DIRA-SS on CIFAR-10C, CIFAR-100C, and ImageNet-C using ResNet architectures under severe common corruptions. 
The results show that DIRA-SS substantially improves performance over the non-adapted source model, achieves accuracy close to the supervised DIRA method, and outperforms existing unsupervised test-time adaptation baselines on ImageNet-C when using only 100 target-domain samples. }

\section{Introduction}
Deep neural network (DNN) classifiers are frequently used in the development of autonomous systems (AS) to enable them to interact and adapt to dynamically changing real-world settings to accomplish their intended goals. 
Using DNNs in autonomous systems has the advantage of producing extremely non-linear decision boundaries to handle the complexity of operating settings since they can learn intricate patterns of complex surroundings. 
However, verifying the behaviour of DNNs to allow for their deployment is challenging. 
Self-driving vehicles are a common example of these ASs. 
According to recent research, an impractical amount of testing is necessary for every software update of a self-driving vehicle to validate the system before it can be put into use~\cite{RR-1478-RC}.
Alternatively, Koopman~et.al.~\cite{Koopman2020}, proposed that after an AS passes a minimum safety validation case, the system is put into use and enhanced during operation to improve reliability.

%
%
%
%

In retraining of DNN classifiers, the optimiser modifies the decision boundary based on the samples used in training. 
During operation, there are usually only a few samples available for retraining.
Retraining using few samples can result in a phenomenon known as \textit{catastrophic forgetting}, where the model fails to generalise to the domain distribution~\cite{Goodfellow2014}. 
%
%
Several methods have been developed, focusing on during operation adaptation of classifiers using few samples both in a supervised and unsupervised (or self-supervised) manner, e.g.~\cite{ghobrial2023dira, mirza2022norm, sun2019unsupervised, sun2020test}. 
State-of-the-art (SOTA) performance of adaptation during operation using few samples has been demonstrated by DIRA (Dynamic Incremental Regularised Adaption)~\cite{ghobrial2023dira}. 
However, DIRA relies on labelled retaining samples.

In the theme of accelerating machine learning, waiting for human operators to provide labels hinders the adoption of such retraining approaches for systems that need to adapt promptly.
Additionally, having a human in the loop to provide labels on demand, creates a cost bottleneck for the utilisation of adaptable machine learning techniques on a large scale.
%
%
In this paper, we discuss the potential of making DIRA self-supervised, to create DIRA-SS.
An experimental evaluation of the success, effectiveness, and limits of DIRA-SS is discussed in this section. 
The rest of this paper is organised as follows, in Section~\ref{sec:related_work} we discuss related work. Section~\ref{sec:method} introduces the method for DIRA-SS. Experimentation setup is explained in section~\ref{sec:experimentation}. Results and discussion handled by section~\ref{sec:results}. Further ablation studies are discussed in section~\ref{sec:ablation}. Lastly, section~\ref{sec:conclusions} concludes and discusses future works.

\begin{figure*}{}
    \centering
    \includegraphics[width=0.75\textwidth]{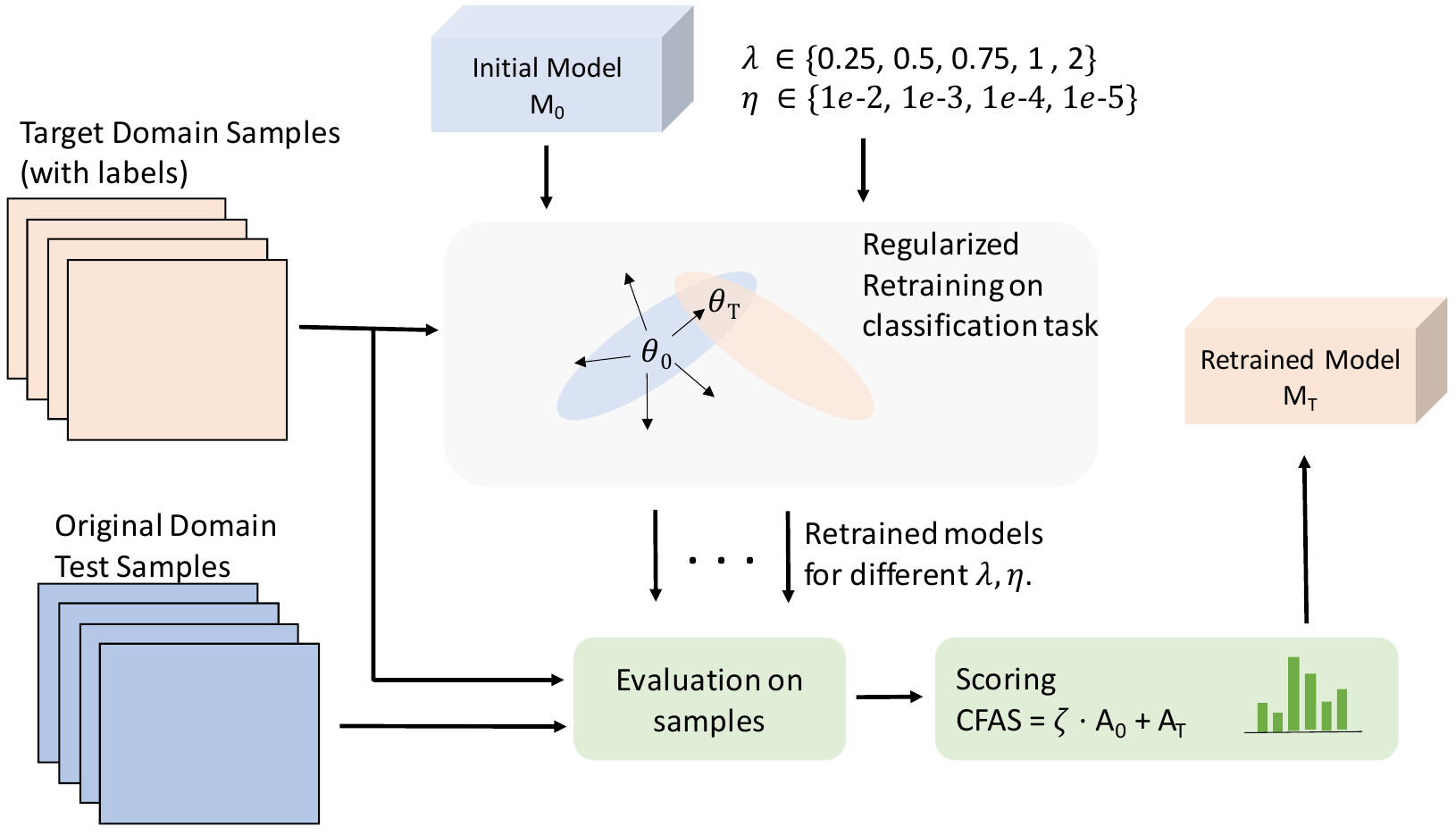}
    \caption{DIRA (Existing Approach)~\cite{ghobrial2023dira}}
    \label{fig:DIRA}
\end{figure*}
{}

{}{}
\section{Related work}\label{sec:related_work}

\subsection{Model generalization}
When training machine learning models, the assumption of independent and identically distributed (i.i.d.) data implies that the training data (source) is drawn from the same distribution as the test data (target) \cite{Zhou2023}\cite{li2019episodic}\cite{li2018deep}\cite{li2017deeper}.
The term `generalisation' is used to refer to a model performing well in a distribution where the i.i.d of data assumption holds.
To formalise this, given a training dataset $D_{train}$ drawn from source distribution $P_{source}$, a model $M$ trained on $D_{train}$ \textit{generalise} to any dataset $D_{test}$ drawn from distribution $P_{source}$.
For many autonomous systems, their operational environments include interacting with an open world, e.g. self-driving vehicles, where the i.i.d assumption is often violated due to the source and target domains exhibiting variations in their data distributions. In other words, the dataset $D_{test}$ would be drawn from a distribution $P_{target}$.
%
%
Hence, the concept of generalisation is difficult to precisely define when the system is expected to encounter environments out of the source data distribution.
Maintaining the i.i.d assumption by rigorously defining the operational environment and having a full understanding of the limitations makes autonomous systems lose their purpose.
The adoption of autonomous systems in the current world is mainly driven by their visioned ability to deal with various scenarios in environments where the possibilities are vast and are impractical to encapsulate during the development of the system.  

Instead, generalisation can be extended to describe a system's ability to adapt to its surrounding operational environment distribution. 
This is a much more scalable approach, goes beyond the assumption of i.i.d data, and opens the venue for systems to operate and generalise in partially defined (or even undefined) environments.
The work introduced in this paper contributes to this latter approach of generalisation whereby machine learning models can adapt to their operational distributions, allowing autonomous systems to fulfil their intended purposes.


%
%

\subsection{Continual learning}
For a formal definition of continual learning and incremental learning please refer to Ghobrial et al.\cite{ghobrial2023dira}.
The concept of continual learning is usually concerned with learning a series of sequential tasks or domains whilst minimising forgetting of previously learnt ones, e.g. \cite{Li2018c}\cite{lopez2017gradient}\cite{Kirkpatrick2017}.
This is done to allow for a model to generalise to a wider set of distributions, through retaining knowledge.
In the case of DIRA-SS, we are approaching the concept of generalisation by equipping models with the ability to quickly adapt and generalise to the current operational environment regardless of the model's performance in other operational environments it operated in previously or will encounter in the future. 
The model will adapt to its current environment dynamically and continuously every time it changes operational domains. 
The retaining of knowledge from previous domains is used by DIRA-SS to utilise transfer learning in order to allow for efficient adaptation using very few samples. 
However, the goal of the model is to maximise its performance on the current domain and drop of performance on previous domains is ignored.


\subsection{One/Few-shot learning}
One-shot or few-shot learning addresses the challenge of training models with limited data to learn a new task.
Unlike traditional machine learning approaches that require training on large datasets to achieve some level of effective performance, one/few-shot learning enables a model to gain good levels of performance from training on a data containing a single or a few instances. This is achieved provided a general representation has been learned by the model previously using a diverse set of samples, e.g.\cite{Song2023}\cite{NEURIPS2022_bcdec1c2}\cite{Schiappa2023}\cite{gidaris2018unsupervised}\cite{gidaris2018dynamic}.
The concept of one/few-shot learning is useful in situations where obtaining extensive data is costly or impractical. 
By utilising methods such as transfer learning one/few-shot learning can equip models with the ability to rapidly adapt to its operational environment to make more accurate predictions. 
The method introduced in this paper, DIRA-SS contributes to the field of one/few-shot self-supervised learning, as it aims to achieve domain adaptation dynamically using few examples that don't have classification labels provided with them.

\subsection{Domain Adaptation Frameworks}
Several approaches have been presented in the literature that address the problem of domain incremental adaptation, e.g.~\cite{sun2020test,sun2019unsupervised,mirza2022norm, schneider2020improving, nado2020evaluating, maria2017autodial, nado2020evaluating, schneider2020improving}.
%
Refer to Mirza et al.~\cite{mirza2022norm} for a brief breakdown of categories of different approaches.
Here we will cover the two SOTA methods relevant to our approach.

The first method is DIRA~\cite{ghobrial2023dira}, which aims at achieving adaptation through the regularisation of old information, whilst retraining on few available samples from the target domain. 
The model is retrained on labelled samples from the target domain. 
DIRA uses elastic weight consolidation~\cite{Kirkpatrick2017} as a regularisation technique to avoid catastrophic forgetting and achieves adaptation using very few samples. 
This allows the model to benefit from transfer learning of information from the initial domain to the target domain.

The second method is TTT (Test-time training)~\cite{sun2020test}\cite{sun2019unsupervised}.
The method combines different self-supervised auxiliary contexts to achieve domain adaptation. 
They break down neural network parameters into three parts, such that pictorially the architecture has a \textit{Y}-structure. 
The bottom section of the Y-structured architecture represents the input layer and the layers responsible for the shared feature extraction, whilst the other two sections contain layers for learning and outputting labels for the main and auxiliary tasks independently. An example of this auxiliary task is being able to tell the rotation of the input image at 90-degree increments i.e.\ 0, 90, 180, 270 degrees. 
During training time the whole neural network is optimised using a combined loss function that aims to maximise performance on both the main and auxiliary tasks. 
During retraining to adapt to a new domain, only parameters of the shared feature extraction and the auxiliary task sections are allowed to change. 
By doing so, the shared feature extraction section of the network modifies to learn the new domain, so that the network may output correct predictions on the unchanged branch of the network responsible for the main task. 
The advantage of TTT is that it is self-supervised because the retraining samples for the auxiliary task can be created by giving a 90-degree incremented rotation to the new domain unlabelled data and the appropriate rotation label is attached. The assumption here is that the new domain input data is always at 0-degree rotation.  
However, the drawback of TTT is that it requires a large number of samples to achieve this self-supervised adaptation.

This paper assesses the feasibility of converting DIRA to a self-supervised technique by using a similar concept of retraining on auxiliary tasks, as is used in the TTT approach. 
However, due to the regularisation available in DIRA, only a few samples are expected to be required for adaptation. Hence, the new method, DIRA-SS, would benefit from the strengths of both DIRA and TTT.
Furthermore, the optimisation problem used by DIRA-SS in the initial training stage is different from that used by TTT. 
TTT tries to optimise on both the main and the auxiliary task at initial training, then in retraining/adaptation, it only optimises on the auxiliary task. 
However, DIRA-SS only optimises on the main task during initial training and only optimises on the auxiliary task during retraining/adaptation.

\section{Method} \label{sec:method}
This section describes the algorithmic details for DIRA-SS.
Let $M_0$ be the model trained on the original domain dataset $X_0$.
The aim of domain adaption is to adapt the trained model parameters, $\theta$, to out-of-distribution target data $X_T$.
The goal of DIRA-SS is to achieve this using a few samples from the target domain and without ground truth labels being provided.
During standard training the optimisation problem tries to solve the loss function $\mathcal{L}$ given by equation~\ref{eq:DIRA_SS_optim_base}.
This usually requires a large number of training samples. 
DIRA~\cite{ghobrial2023dira} modified this optimisation problem and included a regularisation term based on elastic weight consolidation~\cite{Kirkpatrick2017}. 
This is to allow the model being retrained to benefit from transfer learning and thus reduce the number of samples required and improve the accuracy achieved through adaptation.

\begin{equation}
    \displaystyle{\minimize_{\theta}} \mathcal{L}(\theta)
    \label{eq:DIRA_SS_optim_base}
\end{equation}

\begin{equation}
    \displaystyle{\minimize_{\theta}} \mathcal{L}_T(\theta) 
+ \sum_{j}^{}\lambda F_{0,j}(\theta_j - \theta^{*}_{0,j})^2
\label{eq:DIRA_SS-DIRA}
\end{equation}

\begin{figure*}
    \centering
    \includegraphics[width=0.75\textwidth]{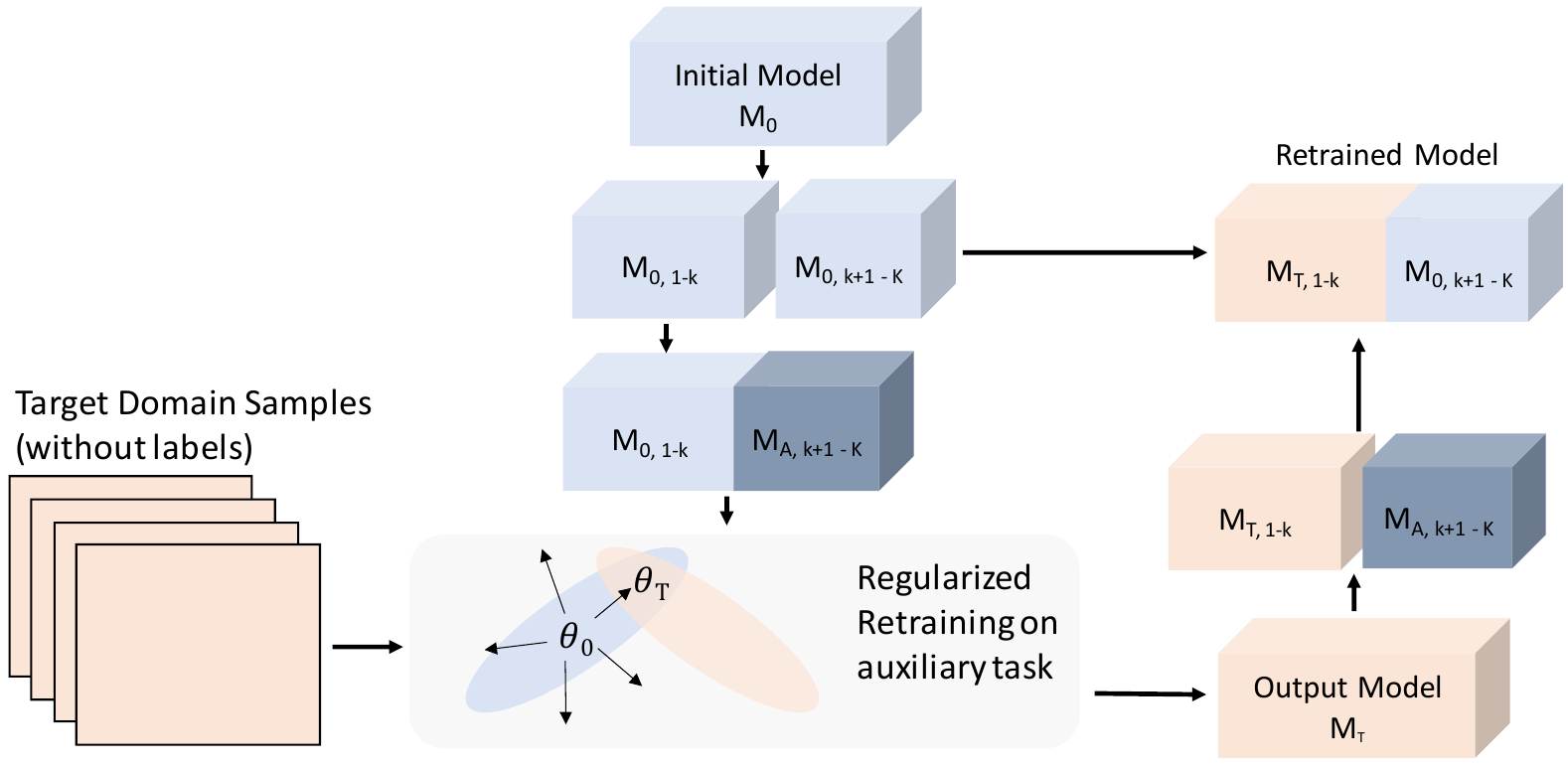}
    \caption{DIRA-SS, the proposed approach combining features of DIRA and TTT.}
    \label{fig:DIRA_SS}
\end{figure*}

The optimisation problem DIRA tries to solve is given by equation~\ref{eq:DIRA_SS-DIRA}, where the loss function for the target domain is represented by the $\mathcal{L}_T$ and  $\sum_{j}^{}\lambda F_{0,j}(\theta_j - \theta^{*}_{0,j})^2$ is an approximation of the original domain implemented based on the method of elastic weight consolidation~\cite{Kirkpatrick2017}.
The approximation term limits how much weights in the model can change, depending on their importance for the original domain. 
Hence benefiting from transfer learning of information from the original domain to learn the target domain.  
The importance of each weight to the information in the original domain is captured in the Fisher matrix, $F_{0,j}$, which is calculated only once when training on the original domain initially. 
This is a step that would be performed during the development phase of a system employing DIRA, i.e.\ it is not calculated during operation when the system is required to adapt to new domains. 
$\lambda$ is a hyper-parameter to allow for fine-tuning to minimise forgetting, and $j$ labels each parameter in the neural network.

The previously developed optimisation problem by DIRA requires labels for the retraining samples to be provided.
The method in this paper, DIRA-SS, aims at extending the optimisation problem proposed by DIRA further to achieve adaptation using a few samples from the target domain and without labels being provided.
The goal is to achieve this by incorporating self-supervision in DIRA.
To achieve self-supervision, the neural network is modified to become a multi-headed model.
One head will be responsible for outputting a classification of the object and the second head is responsible for classifying some auxiliary task.
The auxiliary task head is used in the retraining setting, whilst the classification task is used in the prediction (operational) mode.

In this work we choose the rotation predict task~\cite{gidaris2018unsupervised}, which has been shown to be effective at representation learning~\cite{sun2020test}, in the application of visual recognition, which is the main focus of our work.
Other auxiliary tasks may need to be explored for different applications, such natural language processing (NLP), this is beyond the scope of our current work but is for future work.
The auxiliary task we use rotates the image by 0, 90, 180, 270 degrees and then the model predicts the angle of rotation as a four-way classification problem. 

In the retraining setting, the auxiliary head is attached and the input images are given a random rotation in multiples of 90 degrees. 
Note, that samples available for retraining are assumed to all be at approximately zero degrees rotation. 
Therefore, when a random rotation of 90 degrees is assigned to an image sample, the appropriate label of 0, 90, 180, or 270 can be assigned to the image.  
Retraining on the auxiliary task allows the model's initial layers to adapt to new domains without providing classification labels of objects in images.
This way, samples are allowed to propagate through the neural network and optimise the weights to interpret the new domain better. 
Figure~\ref{fig:DIRA_SS} shows a representation of how we suggest DIRA can be modified to create DIRA self-supervised (DIRA-SS). The letter `A' symbolises relevance to the auxiliary task, `0' to the initial domain, and `T' to the target domain. 
The point at which the multi-head split of the neural network is done can start at any layer $k\in\{1,..., K\}$, where $K$ is the number of layers in the neural network. 

\begin{figure}[]
\centering
    \begin{tikzpicture}[node distance=3.5cm]

         \draw (0,0) -- node[anchor=south, xshift=-1.3cm, yshift =-0.2cm] {$\boldsymbol\theta_{S}$} (1.5,0);
         
         \draw (1.5,0) -- node[anchor=north, xshift=3.3cm, yshift =1cm, align=center] {$\boldsymbol\theta_{R}$ (auxilary task branch /\\ retraining branch)} (3,1);

         \draw (1.5,0) -- node[anchor=south, xshift=3.3cm, yshift =-1cm, align=center] {$\boldsymbol\theta_{M}$ (main task branch /\\prediction branch)} (3,-1);

    \end{tikzpicture}
\caption{DIRA-SS model parameters split schematic.}
\label{fig:y_shaped_model}
\end{figure}

To formalise DIRA-SS, the main and auxiliary tasks share some of the model parameters $\boldsymbol\theta_{S} = (\theta_1, ..., \theta_k)$ which include parameters in the input layer up until layer $k \in \{ 1, ..., K\}$. 
The main task branch $\boldsymbol\theta_{M} = (\theta_{k+1}, ..., \theta_K)$ contains the rest of the pre-trained layers used in predicting the main task.
The auxiliary task gets its own branch with another set of parameters $\boldsymbol\theta_{R} = (\theta^{'}_{k+1}, ..., \theta^{'}_K)$, this is the \textit{retraining} branch. 
See figure~\ref{fig:y_shaped_model} for a pictorial description of the joint architecture. 
For the retraining and predictions branches, they have the same architecture except for the output layer, due to the different number of classes between the main and the auxiliary tasks.
%

Initial training is done for model parameters $\theta_S$ and $\theta_M$ on the main task using data samples from the original domain, to create the initial model $M_0$.
After training on the main classification task, the auxiliary branch is then added to the model. The auxiliary branch, comprising of weights $\theta_R$, is allowed to fine-tune on the auxiliary task whilst all other layers in the model are frozen. Note, this allows our method to be integrated with any existing single-headed pretrained model.

%
During domain adaptation, the model parameters $\theta_S$ and $\theta_R$ are allowed to optimise on the auxiliary task using samples from the target domain.
This is done whilst parameters $\theta_S$ are regularised on the original domain using elastic weight consolidation as was done in DIRA, to achieve adaptation using a few retraining samples.
The optimisation problem therefore for DIRA-SS can be formalised as shown by equation~\ref{eq:DIRA-SS}.

\begin{equation}
    \displaystyle{\minimize_{\theta_S,\theta_R}} \mathcal{L}_A(\theta_S, \theta_R) 
+ \sum_{j}^{}\lambda F_{0,j}(\theta_{S,j} - \theta^{*}_{0,j})^2
\label{eq:DIRA-SS}
\end{equation}

To utilise elastic weight consolidation during retraining there are certain assumptions made in DIRA. 
Since DIRA-SS utilises elastic weight consolidation as well, these assumptions need to be maintained. 
Satisfying these assumptions involves two conditions when retraining.
The first condition, computing the Fisher matrix requires the distribution to be well expressed using a large pool of samples from the target domain.
In the case of DIRA-SS and DIRA, this large pool of samples from the target domain is not available.  
Therefore, the Fisher matrix used in the optimisation problem is calculated using the original training dataset during initial training and a copy of this calculated Fisher matrix is stored for retraining, omitting the need to keep initial training data.
The assumption that comes with the Fisher matrix, is that retraining starts from a model that is already optimised on the dataset represented by the Fisher matrix.
This leads to the second condition.
Since the system only has access to the Fisher matrix from the original domain, retraining using any number of samples, will need to start from the original model $M_0$.
Practically this is achievable as a copy of $M_0$ can always be kept onboard of a system.
This is to maintain the validity of the Fisher matrix used in the optimisation problem. 
For more elaboration on these conditions, refer to the Ghobrial et al.~\cite{ghobrial2023dira}.
%
%

%

In each training step $t$, the shared feature extractor represented by model parameters $\theta_S$ is updated according to Equation~\ref{theta_new_substituted}, where $\eta$ is the learning rate. 
\begin{equation}
   \theta_{S,t+1} = \theta_{S,t} - \eta \bigg( \dfrac{\mathcal{L}_{\text{A}}(\theta_S)}{d\theta} - 
   \dfrac{\sum_{j}^{}\lambda F_0(\theta_{S,t,j} - \theta^{*}_{0,j})^2}{d\theta} \bigg)
   \label{theta_new_substituted}
\end{equation}

Similar to DIRA, the two hyperparameters, $\eta$ and $\lambda$, are critical for the success of DIRA-SS. 
%
%
%
From empirical testing, done in DIRA it was found that a combination of $\eta = 1$e-5 and $\lambda=1$ yields near optimum adaptation.
These values were maintained for DIRA-SS too.

\section{Experimentation Setup}\label{sec:experimentation}

We used the problem of image classification to showcase DIRA-SS. 
All of our experimentation was based in PyTorch library~\cite{paszke2019pytorch}. 
This section discusses the experimentation setup details used in evaluating DIRA-SS. Code is available at this repository: \url{https://github.com/Abanoub-G/DIRA-SS}

\subsection{Benchmarks}
We utilise CIFAR-10C, CIFAR-100C, and ImageNet-C datasets in our experimentation. These are image classification dataset benchmarks. They are aimed at evaluating the robustness of models against common corruptions~\cite{hendrycks2019benchmarking}.
The benchmarking datasets introduce different corruptions to the test sets of CIFAR-10/CIFAR-100\cite{krizhevsky2009learning} and ImageNet~\cite{deng2009imagenet}
There are 20 types of corruptions in total. For each corruption, five different levels of severities are provided. 
Most state-of-the-art (SOTA) domain-incremental retraining frameworks utilise 15 corruptions out of the 20 in their comparisons, e.g.~\cite{mirza2022norm, sun2020test}. 
These are deemed the more common corruptions. 
We use the same 15 common corruptions used by other methods in the literature.


\subsection{Baselines}
We list below the different baselines we assess against our DIRA-SS approach:
\begin{enumerate}
    \item \textbf{Source}: Refers to results of the corresponding baseline model trained on the incorrupt data, without adaption to the target domain.


    \item \textbf{TTT}~\cite{sun2020test}: Test-Time Training (TTT) adapts parameters in the initial layers of the network by using auxiliary tasks to achieve self-supervised domain adaption. 
    
    \item \textbf{NORM}~\cite{schneider2020improving, nado2020evaluating}: Ignores the initial training statistics and recalculates the batch normalization statistics using samples from the target domain only (requiring a large number of samples).

    \item \textbf{DUA}~\cite{mirza2022norm}: Dynamic Unsupervised Adaptation (DUA), takes into account initial training statistics and updates batch normalization statistics using samples from the target domain (requiring few samples).

    \item \textbf{DIRA}~\cite{ghobrial2023dira}: Dynamic Incremental Regularised Adaptation (DIRA), is a method that used regularisation to benefit from transfer learning when adpating this is the method that DIRA-SS extends to make it unsupervised. DIRA is a supervised method, but required few samples).
    
\end{enumerate}

\subsection{Models and Hardware}
We used ResNets~\cite{he2016deep} in our experiments, utilising two versions of ResNet: ResNet-18 (18-layer) and ResNet-26 (26-layer). 
For CIFAR-10/CIFAR-100, we used ResNet-26 as this is the version of ResNet architecture optimised for CIFAR-10/CIFAR-100 and is used by others in the literature. 
Initial training for the models was done locally. 
For ImageNet, we used a pre-trained off-the-shelf ResNet-18 model from PyTorch. 
%
%
%
Experiments for CIFAR10 and CIFAR100 were done on an MSI GF65 THIN 3060 Laptop with 64GB RAM and a Linux Ubuntu 20.04.2 LTS (64-bit) operating system, whilst for ImageNet a Dell Alienware Desktop PC with 64GB RAM and a Linux Ubuntu 18.04.4 LTS (64-bit) operating system were used.

To achieve reliable comparisons against baselines the starting model parameters from which retraining is done must be the same.
Otherwise, the accuracy improvements cannot be reliably attributed to the effectiveness of the retraining method and can be argued that it is due to varying starting model accuracies or paramters.
Therefore, in our results for CIFAR10/100 on ResNet-26 we only compare against Source, as we do not have the initial models used in retraining by other SOTA methods.
For ImageNet on ResNet-18 we compare against SOTA methods because the starting trained model used by other SOTA retraining methods is the same off-the-shelf ResNet-18 model from PyTorch.
This is a controlled experiment with same exact model initialisation and initial training model parameters, and therefore yields a fair and honest comparison between the methods.

\subsection{Optimisation Details}
Stochastic Gradient Decent (SGD) was used for optimisation during training and retraining.
For an initially trained model, its architecture gets extended to include the auxiliary task. 
The model is then allowed to fine tune the auxiliary task layers only, whilst the rest of the neural network is frozen, using two epochs of training on the auxiliary task and the initial training dataset.
This creates the initial model $M_0$ that is then used in retraining, when adaptation is required.

During adaption to the target domain we used a learning rate value of $\eta = $1$e$-5 and a lambda value of $\lambda = 1$. 
Empirically these were found to be suitable hyperparameters for optimised retraining. 
%
%
%
%
The retraining is quick as only a small number of samples are used and the retraining is done over 10 epochs.
%
%
%
We use top-1 classification accuracy as our assessment metric in all experiments~\cite{ghobrial2023evaluation}.

\section{Results \& Discussion}
\label{sec:results}

\subsection{Initial analysis on layers significance}
An initial analysis was conducted on the layers that get adapted the most during retraining using DIRA. 
This is done to generate an initial understanding of whether DIRA-SS can potentially be successful. 
The general idea here is that if the initial layers are the ones that adapt the most, then adapting to representations learnt from the auxiliary task will be beneficial for successful adaptation.
On the other hand, if the last layers (layers closer to the output) need to adapt the most then the benefit from adaptation using the auxiliary task may be limited.

\begin{figure}[h]
    \centering
    \includegraphics[width=0.5\textwidth]{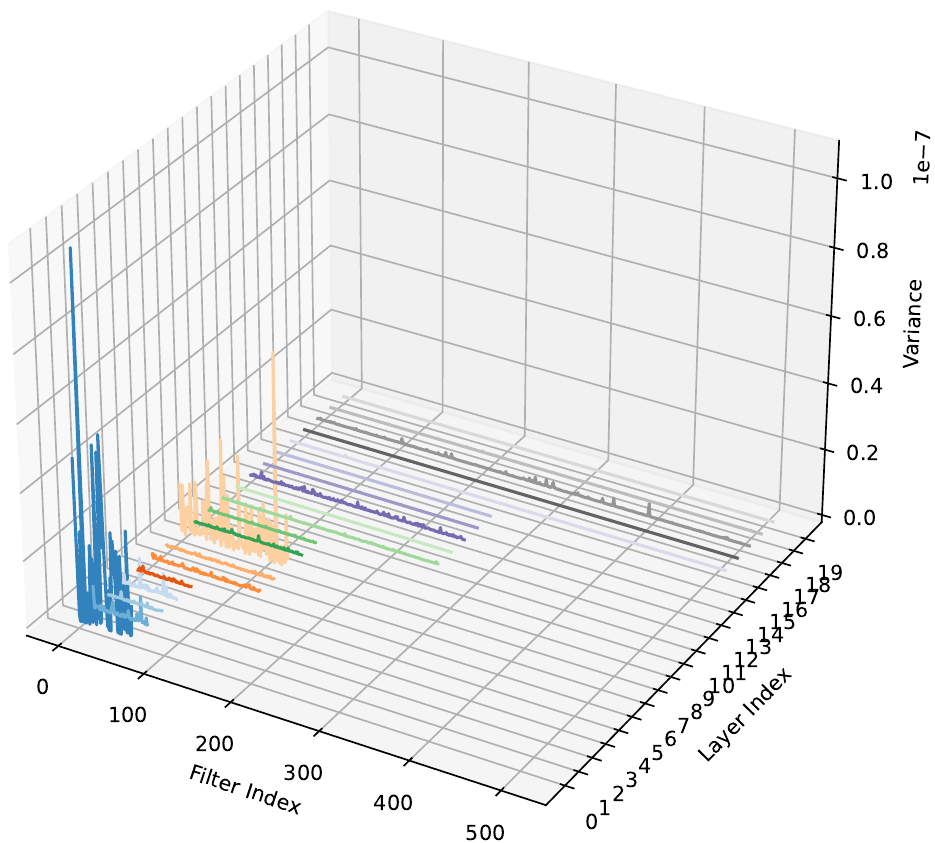}
    \caption{3D plot showing variances in ResNet-18 layers across different retrained models on CIFAR10C~\cite{hendrycks2019benchmarking} using DIRA~\cite{ghobrial2023dira}.}
    \label{fig:Variances_3D}
\end{figure}

To do this analysis the variance for model parameters between different adaptations of a ResNet-18 model using DIRA to different domains (noise types) in the CIFAR-10C benchmarking dataset, using different numbers of samples from 0 to 100 is calculated. 
Figure~\ref{fig:Variances_3D} shows a 3D visualisation of the variance for the model parameters.  
The layer index enumeration starts from the input side of the neural network and increases to reach the output side of the network.
Layers closer to the input side seem to vary more compared to layers closer to the output side, in particular there are two layers showing the highest variance.
This shows affirming indications that retraining using the auxiliary task will result in beneficial representations being learnt by the model.
Investigating which representations are learnt by these layers and the exact reasons why specifically these two layers mount to the most effect on adaption is beyond the scope of this work and can be of interest to future work.

\subsection{Overall Adaptation}
\begin{figure}[h]
    \centering
    \includegraphics[width=0.5\textwidth]{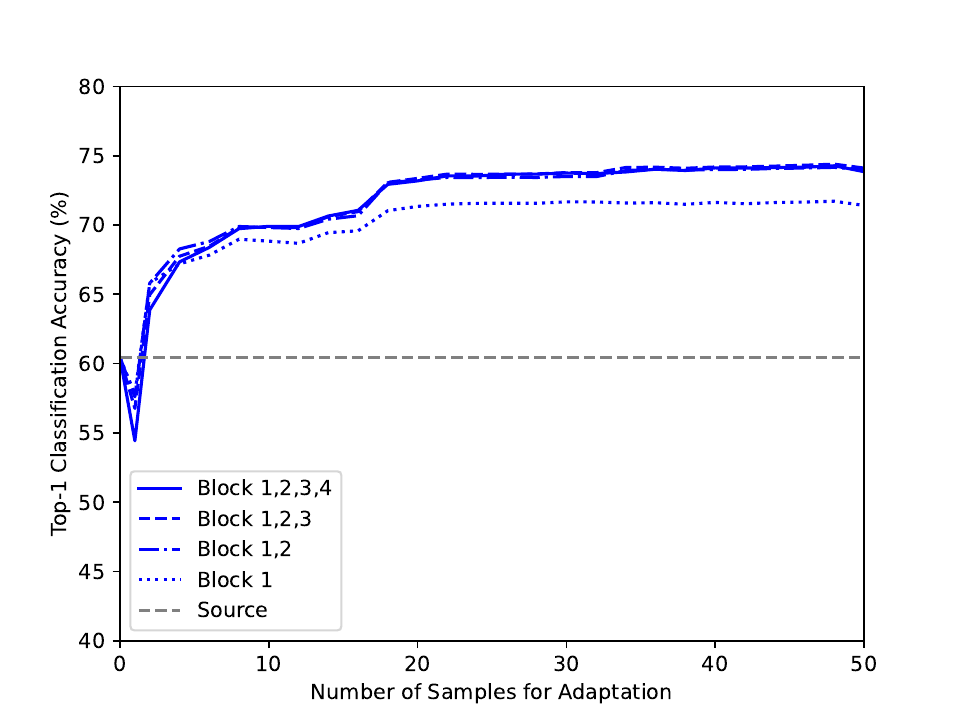}
    \caption{ResNet-26 mean classification accuracy over 15 different corruption types on CIFAR-10C at the highest severity (Level 5). Plot shows adaptation using DIRA-SS for different split points in the ResNet-26 model i.e. increasing the shared model parameters, $\theta_S$, from only a couple of initial layers all the way to sharing the whole model apart from the output layer. Blocks are used, to refer to the different sections of a ResNet, which consists of four blocks.}
    \label{fig:layers}
\end{figure}

Figure~\ref{fig:layers} shows how top-1 classification accuracy changes as the number of samples available from the target domain increases.
The plot shows the effect of having different portions of the model parameters being shared between the auxiliary and the main tasks.
ResNets, which are used in this study, consist of four blocks of layers. 
The adaptation is done using four different settings, progressively adding more layers to the shared set of model parameters in each setting. 
Visualising the results through two perspectives: 1) in the presence of approximately five data samples, and 2) when the data samples exceed five.
%
For the first perspective, adapting all blocks results in the least effective adaptation out of the four settings, whilst adapting only the first and second blocks results in the best outcome.
For the second perspective, adapting only the first block of layers results in the worst outcome, whilst all other three settings approximately result in the same outcome.
From this analysis, it can be deduced that sharing the first and the second blocks results in the most optimum adaptation setting for DIRA-SS.
This also coincides with observations from Figure~\ref{fig:Variances_3D}, which shows that for good adaptation to new domains, the model needs to mainly adapt in its initial layers which fall in blocks 1 and 2 for ResNets.

To analyse the success of DIRA-SS method further, Table~\ref{tab:DIRA-SS_CIFAR-10C_results} and Table~\ref{tab:DIRA-SS_CIFAR-100C_results}, compares between DIRA and DIRA-SS performance on the fifteen types of common corruptions from the benchmark datasets used for CIFAR-10 and CIFAR-100, respectively. 
Results are shown for retraining using 100 samples. 
In both methods, the adaption improves significantly compared to Source.
DIRA achieves overall better adaptation than DIRA-SS. 
The improvement DIRA provides over DIRA-SS, however, is very little (less than $1\%$).
This is impressive given that DIRA-SS is self-supervised/unsupervised, whilst DIRA is a supervised method. 
%

%

\begin{table*}[]
    \centering
    \resizebox{\textwidth}{!}{\begin{tabular}{c|c c c c c c c c c c c c c c c|c}
                 & gaus & shot & impul & defcs & gls & mtn & zm & snw & frst & fg & brt & cnt & els & px &jpg & mean \\\hline
Source & 58.5 & 61.3 & 37.3 & 51.9 & 59.6 & 58.6 & 58.1 & 73.3 & 67.8 & 50.0 & 80.7 & 19.2 & 71.8 & 66.1 & \textbf{79.8} & 59.6 \\
DIRA  & \textbf{73.6} & \textbf{75.6} & \textbf{61.9} & 79.7 & \textbf{65.8} & \textbf{77.9} & \textbf{80.0} & \textbf{77.4} & \textbf{77.0} & 72.6 & \textbf{84.2} & 60.2 & \textbf{74.9} & \textbf{76.9} & 79.5 & \textbf{74.5} \\ 
DIRA-SS  & 72.8  & 74.2 & 60.0 & \textbf{81.0} & 62.2 & 76.7 & 77.4 & 75.9 & 74.4 & \textbf{73.9} & 83.0 & \textbf{73.6} & 72.4 & 76.4 & 77.3 & 74.1 \\ 

    \end{tabular}}
    \caption{Top-1 Classification Accuracy (\%) for each corruption in CIFAR-10C at the highest severity (Level 5). Source shows the results from the same model trained on the clean train set (CIFAR-10) and tested on the corrupted test set (CIFAR-10C). ResNet-26 is used. Highest accuracy is highlighted in bold.}
    \label{tab:DIRA-SS_CIFAR-10C_results}
\end{table*}

\begin{table*}[]
    \centering
    \resizebox{\textwidth}{!}{\begin{tabular}{c|c c c c c c c c c c c c c c c|c}
                 & gaus & shot & impul & defcs & gls & mtn & zm & snw & frst & fg & brt & cnt & els & px &jpg & mean \\\hline
Source  & 24.2 & 27.0 & 9.7 & 30.0 & 30.9 & 33.6 & 35.5 & 38.8 & 34.6 & 19.6 & 44.8 & 8.4 & 43.4 & 39.8 & 50.0 & 31.4 \\
DIRA  & \textbf{44.7} & 45.1 & \textbf{33.6} & 50.9 & \textbf{40.4 }& \textbf{49.6} & \textbf{52.3} & \textbf{47.3} & \textbf{46.6} & 37.9 & \textbf{55.2} & 33.3 & \textbf{47.0} & \textbf{51.5} & \textbf{51.7} & \textbf{45.8} \\ 
DIRA-SS  & 43.2 & \textbf{45.6} & 31.8 & \textbf{53.1} & 39.3 & 45.5 & 51.8 & 43.2 & 45.3 & \textbf{40.9} & 53.7 & \textbf{39.5} & 45.8 & 49.9 & 49.6 & 45.2 \\ 

    \end{tabular}}
    \caption{Top-1 Classification Accuracy (\%) for each corruption in CIFAR-100C at the highest severity (Level 5). Source shows the results from the same model trained on the clean train set (CIFAR-100) and tested on the corrupted test set (CIFAR-100C). ResNet-26 is used. Highest accuracy is highlighted in bold.}
    \label{tab:DIRA-SS_CIFAR-100C_results}
\end{table*}

\begin{table*}[]
    \centering
    \resizebox{\textwidth}{!}{\begin{tabular}{c|c c c c c c c c c c c c c c c|c}
                 & gaus & shot & impul & defcs & gls & mtn & zm & snw & frst & fg & brt & cnt & els & px &jpg & mean \\\hline
Source  & 1.6 & 2.3 & 1.6 & 9.4 & 6.6 & 10.2 & 18.2 & 10.5 & 15.0 & 13.7 & 48.9 & 2.8 & 14.7 & 23.1 & 28.3 & 13.8\\
TTT  & 3.1 & 4.5 & 3.5 & 10.1 & 6.8 & 13.5 & 18.5 & 17.1 & 17.9 & 20.0 & 47.0 & \textbf{14.4} & 20.9 & 22.8 & 25.3 & 16.4\\
NORM  & \textbf{12.9} & 10.4 & 9.5 & \textbf{12.4} & 10.6 & \textbf{20.0} & 28.1 & \textbf{29.4} & 18.5 & 33.1 & 52.2 & 10.2 & 26.5 & 35.8 & 31.5 & 22.7\\
DUA  & 10.6 & \textbf{12.4} & \textbf{11.9} & 12.0 & \textbf{11.4} & 15.3 & 25.7 & 22.2 & 21.6 & 31.4 & 54.4 & 4.1 & 27.8 & 33.5 & \textbf{32.6} & 21.8\\
DIRA-SS  & 11.1 & 11.8 & 11.2 & 9.6 & 10.4 & 16.4 & \textbf{30.8} & 26.2 & \textbf{26.6} & \textbf{35.9} & \textbf{56.0} & 7.5 & \textbf{35.7} & \textbf{39.7} & 29.5 & \textbf{23.9}\\ \hline
DIRA  & 12.0 & 13.5 & 11.6 & 10.2 & 11.5 & 18.7 & 31.2 & 26.6 & 27.2 & 36.3 & 56.3 & 9.2 & 35.7 & 38.1 & 32.0 & 24.7\\ 
    \end{tabular}}
    \caption{Top-1 Classification Accuracy (\%) for each corruption in ImageNet-C at the highest severity (Level 5). Source refers to results from the same model trained on the uncorrupted training dataset (ImageNet) and tested on the corrupted test dataset (ImageNet-C). A fair comparison is maintained with TTT, NORM, DUA, and DIRA by using the same initially trained ResNet-18 model with the same training checkpoint and supplying the same number of retraining samples. The values for the performance from the other frameworks (TTT, NORM, DUA, DIRA) were used from~\cite{ghobrial2023dira}. 100 retraining samples are used. Highest accuracy is highlighted in bold.
    }
    \label{tab:DIRA-SS_SOTA_ImageNet_results}
\end{table*}

\subsection{Dynamic adaptation scenario for DIRA-SS}
\begin{figure}[h]
    \centering
    \includegraphics[width=0.5\textwidth]{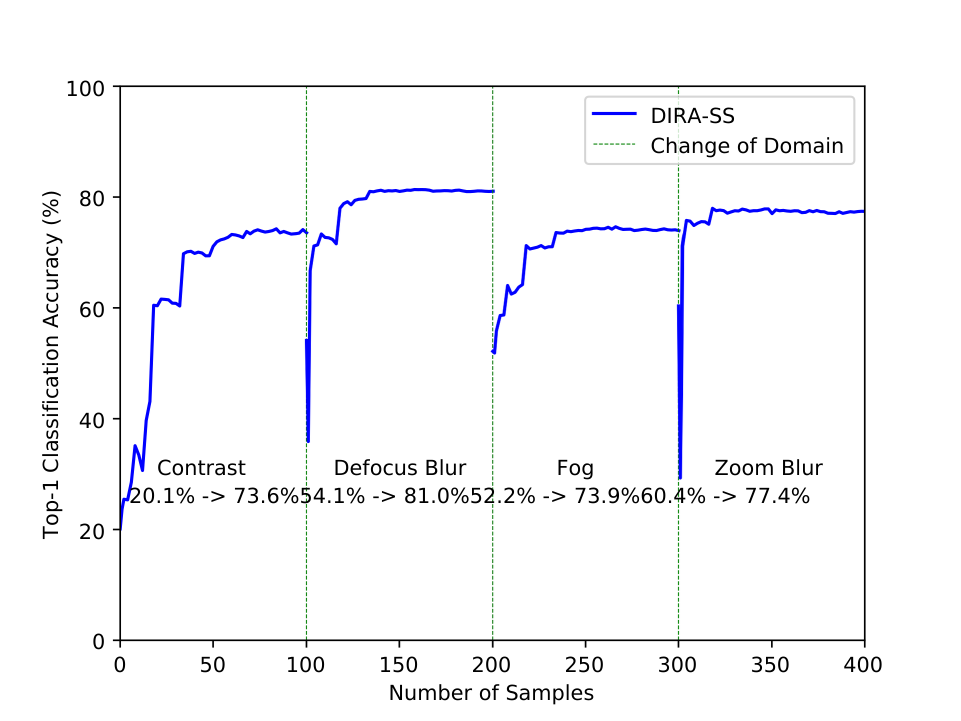}
    \caption{Dynamic adaptation scenario example for DIRA-SS to four different domains from CIFAR-10C. Pre-trained ResNet-26 on CIFAR-10 adapts to different corruption examples from CIFAR-10C dataset at the highest severity (Level 5), to show how well DIRA-SS can dynamically adapt to operational domains.}
    \label{fig:dira_ss_dynamic}
\end{figure}

\begin{figure*}[h]
    \centering
    \includegraphics[width=1\textwidth]{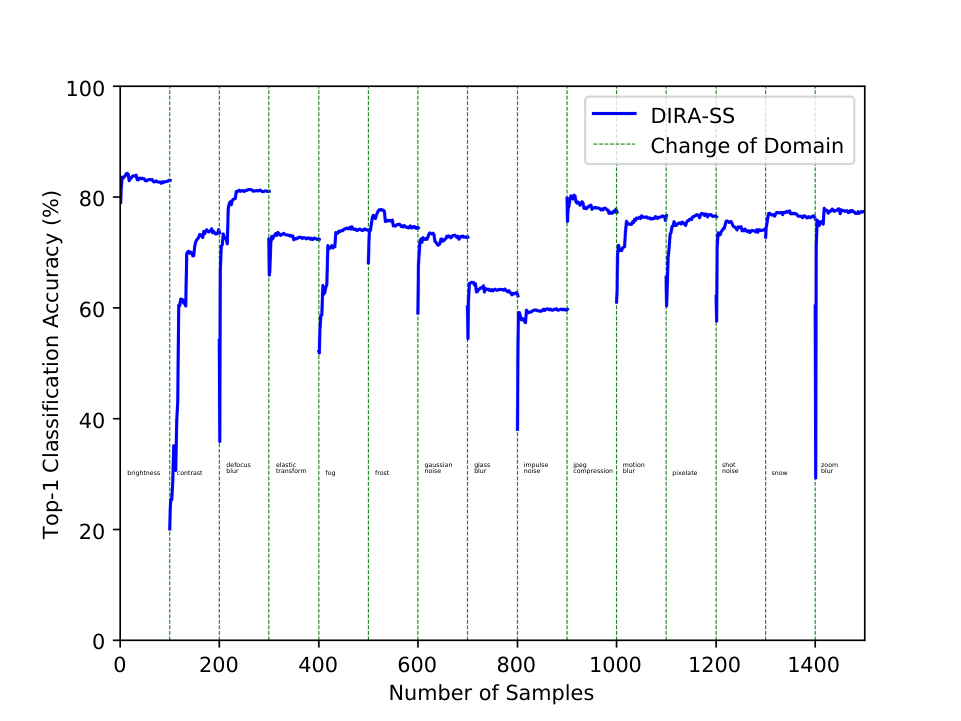}
    \caption{Dynamic adaptation scenario example for DIRA-SS to all different domains from CIFAR-10C. Pre-trained ResNet-26 on CIFAR-10 adapts to different corruption examples from CIFAR-10C dataset at the highest severity (Level 5), to show how well DIRA-SS can dynamically adapt to operational domains.}
    \label{fig:dira_ss_dynamic_all}
\end{figure*}

In real-life scenarios, varying domains may occur during operation. 
To visualise how DIRA-SS adapts to continuously changing domains, Figure~\ref{fig:dira_ss_dynamic} is plotted to show the shift to four different domains consecutively from CIFAR-10C. 
The results depicted show that the accuracy improves as samples from the target domain increase to a certain point then plateaus. 
The number of samples at which the plateauing happens varies depending on the noise type, but by 100 samples a plateau is reached for the four different types of domains used.
Furthermore, Figure~\ref{fig:dira_ss_dynamic_all} shows the adaptation to the fifteen types of domains available in the benchmarking dataset for CIFAR-10C.

\subsection{Comparison with SOTA}
To assess how well our approach performs compared with SOTA domain adaptation frameworks we compare results with three domain adaptation frameworks from the literature: TTT, NORM and DUA, on the ImageNet dataset. 
Table \ref{tab:DIRA-SS_SOTA_ImageNet_results} show top-1 classification accuracy for the highest severity level on dataset ImageNet-C.
The DIRA-SS method performs competitively against SOTA domain unsupervised adaptation approaches. 
As can be seen from the table, it achieves SOTA overall performance averaged between the different corruptions in the benchmarking datasets. 
This is while using a limited number of samples from the target domain (100 samples).
DIRA-SS overall performance drops slightly compared to DIRA, however, this small trade-off is expected as one drops the supervised learning provided by DIRA to adopt the self-supervision offered by DIRA-SS.

\subsection{Retraining Time}
\begin{figure}[]
    \centering
    \includegraphics[width=0.5\textwidth]{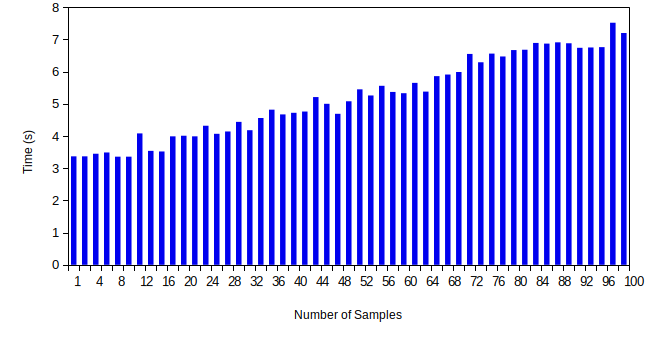}
    \caption{Time consumption for DIRA-SS retraining ResNet-26 on a different number of samples of CIFAR-10C dataset.}
    \label{fig:time}
\end{figure}

Retraining time is a crucial aspect of runtime adaption. 
Specifically in autonomous systems that operate in time or safety-critical applications.
Therefore, studying retraining time is very important.
Figure~\ref{fig:time} shows the time consumption for retraining ResNet-26 using a different number of samples on CIFAR-10C with DIRA-SS.
The time consumption increases as the number of samples increases as one would expect. 
For this particular case study, the time consumption seems to range between 3 to 6+ seconds. 
Whilst this is relatively fast compared to initial training, it may still not be fast enough for applications where near instantaneous update of models to their operational domains is required i.e. retraining may need to be completed in the matter of milli- or micro-seconds. 

Time consumed in retraining can be reduced by making models more efficient, as autonomous systems are expected to adapt in highly dynamic environments where speed is a critical factor for safety.  
This can be achieved through neural network compression, e.g.~\cite{ghobrial2023evaluation}, optimised to run on hardware.
Extending this work to explore the adaptation of compressed models on limited-resource hardware, is a topic of interest and is left for future work.

Furthermore, energy consumption is another topic of interest falling in the same context of efficient models that operate on limited-resources hardware.
Making retraining methods more energy efficient can also be useful for applications with limited energy budgets, e.g. space applications.

\section{Ablation Studies}
\label{sec:ablation}
\subsection{Samples Type Effect}
\begin{figure}[]
    \centering
    \includegraphics[width=0.5\textwidth]{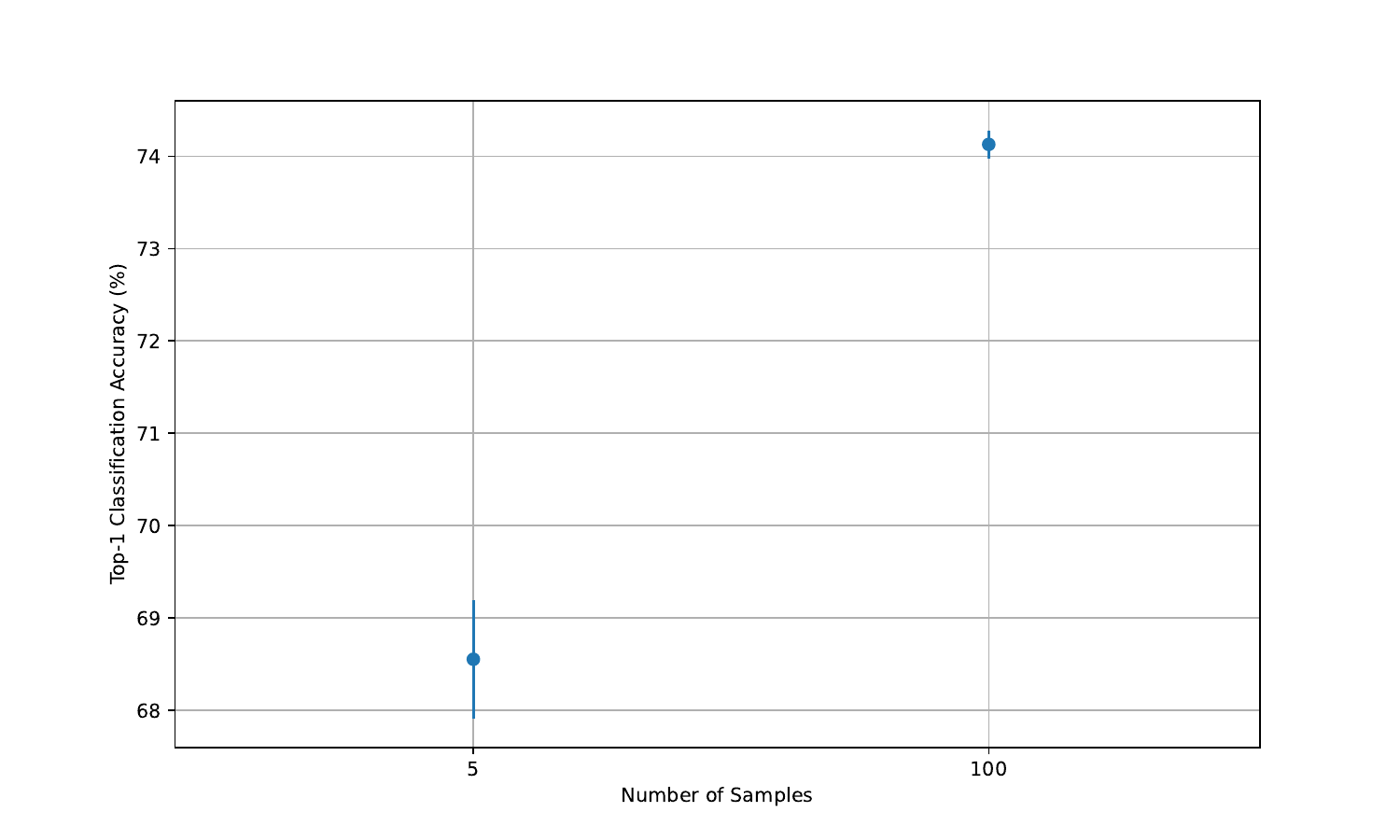}
    \caption{ResNet-26 model Top-1 accuracy after retraining using DIRA-SS on different combinations of 5 and 100 samples from CIFAR-10C dataset.}
    \label{fig:samples_effect_error_DIRA_SS}
\end{figure}

In all of the previous experiments, seeds were fixed to ensure the same samples were drawn and in the same order for all experiments.  
This was done to maintain a fair comparison between experiments and avoid any observed improvements in results happening due to uncontrolled experiment environments. 
In this section, however, the effect of using different samples, regardless of their order, is studied to generate a better understanding of the effect of changing samples used in retraining.
To do this, the fixing of seeds is removed and retraining is run for 5 and 100 samples.  
The experiment is repeated two hundred times, and the mean and standard deviation are calculated. 
The source accuracy of the model is $59.6\%$
Figure~\ref{fig:samples_effect_error_DIRA_SS} plots the error bar results for the experiment. 
The larger standard deviation over the 200 runs occurs at adaptation on 5 samples, where it achieves $68.6\pm 0.7$.
At 100 samples the model achieves $74.1\pm 0.1$
As can be seen, the effect of which samples are used and the order is minimal on the improved performance of the model from retraining on DIRA-SS.

\subsection{Effect of Augmentation}
In works, such as DUA~\cite{mirza2022norm}, it was shown that using data augmentation had benefits to their retraining methods performance.
%
%
Different types of augmentations exist, for example, rotate, vertical-flip, horizontal-flip, crop, etc. 
%
%
Throughout DIRA-SS experiments shown in this paper data augmentations were not used, as it was thought that it might have a negative impact on the auxiliary task. 
The auxiliary task can be seen as a form of data augmentation in itself because it incorporates rotations into the retraining.
Due to the orientation being a building block in driving the self-supervised representation learning for DIRA-SS, data augmentations were avoided.

\section{Conclusions and Future Works} \label{sec:conclusions}

This paper introduced DIRA-SS (Dynamic Incremental Regularised Adaptation - Self Supervised), a retraining method that allows for online domain adaption using few samples from the target domain. The method in essence advances the previously introduced method DIRA to make it unsupervised instead of supervised, by using the concept of auxiliary tasks to achieve self-supervised retraining. This is achieved by adding an additional head to the existing model. The original model head is used in classification. The additional head is used in retraining to allow the initial layers of the model to adapt without affecting the last layers in the model. From analysis carried out, it was found that initial layers in the model are the ones that adapt the most in domain adaptation. 
The auxiliary task used in DIRA-SS is rotation classification. Assuming all images input into the model during operation are at 0 degrees orientation, one can create augmented orientations of the image and allocate the rotation label to it, thus achieving self-supervised retraining.
The method has been evaluated on benchmarks used widely to assess methods for domain adaption. The findings of this paper support the concept of using auxiliary tasks to achieve unsupervised or self-supervised online adaptation of neural network models. DIRA-SS has proven to be able to still achieve effective domain adaption without the need for providing classification labels during retraining. A trade-off in accuracy is experienced by DIRA-SS due to the absence of labels, in comparison with DIRA, this trade-off is typically less than 1\% on average. However, DIRA-SS still achieves higher performance than other unsupervised state-of-the-art methods in the literature.

For future works, the retraining method introduced in this paper achieve state-of-the-art domain adaptation accuracies, further advancements are still required to achieve higher adaptation accuracies to reach the necessary levels required for reliable continuous adaption. The paper thesis did not explore thoroughly the retraining effects on neural network layers. An in-depth understanding of representations learned by different layers and the precise effect behind the adaptation of specific layers can perhaps lead to enhanced adaptability performance.
Another director for future work is getting adaptation methods to function on real systems where resources are limited and speed of retraining is key to the adoption of a method requires more research. A future direction involves the exploration of retraining methods on compressed models deployed on limited-resource hardware or edge devices.

\section*{Acknowledgments}
This research is part of an iCASE PhD funded by EPSRC and Thales UK. 


\printbibliography

@inproceedings{mirza2022norm,
  title={The norm must go on: dynamic unsupervised domain adaptation by normalization},
  author={Mirza, M Jehanzeb and Micorek, Jakub and Possegger, Horst and Bischof, Horst},
  booktitle={Proceedings of the IEEE/CVF Conference on Computer Vision and Pattern Recognition},
  pages={14765--14775},
  year={2022}
}

@article{Kirkpatrick2017,
archivePrefix = {arXiv},
arxivId = {1612.00796},
author = {Kirkpatrick, James and Pascanu, Razvan and Rabinowitz, Neil and Veness, Joel and Desjardins, Guillaume and Rusu, Andrei A. and Milan, Kieran and Quan, John and Ramalho, Tiago and Grabska-Barwinska, Agnieszka and Hassabis, Demis and Clopath, Claudia and Kumaran, Dharshan and Hadsell, Raia},
doi = {10.1073/pnas.1611835114},
eprint = {1612.00796},
issn = {10916490},
journal = {Proceedings of the National Academy of Sciences of the United States of America},
number = {13},
pages = {3521--3526},
pmid = {28292907},
title = {{Overcoming catastrophic forgetting in neural networks}},
volume = {114},
year = {2017}
}

@article{Li2018c,
archivePrefix = {arXiv},
arxivId = {1606.09282},
author = {Li, Zhizhong and Hoiem, Derek},
doi = {10.1109/TPAMI.2017.2773081},
eprint = {1606.09282},
issn = {19393539},
journal = {IEEE Transactions on Pattern Analysis and Machine Intelligence},
number = {12},
pages = {2935--2947},
pmid = {29990101},
publisher = {IEEE},
title = {{Learning without Forgetting}},
volume = {40},
year = {2018}
}

@book{RR-1478-RC,
author="Kalra, Nidhi and Susan M. Paddock",
title="Driving to Safety: How Many Miles of Driving Would It Take to Demonstrate Autonomous Vehicle Reliability?",
address="Santa Monica, CA",
year="2016",
doi="10.7249/RR1478",
publisher="RAND Corporation"
}

@article{Goodfellow2014,
archivePrefix = {arXiv},
arxivId = {1312.6211},
author = {Goodfellow, Ian J. and Mirza, Mehdi and Xiao, Da and Courville, Aaron and Bengio, Yoshua},
eprint = {1312.6211},
journal = {2nd International Conference on Learning Representations, ICLR 2014 - Conference Track Proceedings},
title = {{An empirical investigation of catastrophic forgetting in gradient-based neural networks}},
year = {2014}
}

@book{Koopman2020,
author = {Koopman, Philip and Wagner, Michael},
doi = {10.1007/978-3-030-55583-2_26},
isbn = {9783030555832},
pages = {351--357},
publisher = {Springer International Publishing},
title = {{Positive Trust Balance for Self-driving Car Deployment}},
url = {http://dx.doi.org/10.1007/978-3-030-55583-2_26},
volume = {2},
year = {2020}
}

@article{hendrycks2019benchmarking,
  title={Benchmarking neural network robustness to common corruptions and perturbations},
  author={Hendrycks, Dan and Dietterich, Thomas},
  journal={ICLR},
  year={2019}
}

@inproceedings{sun2020test,
  title={Test-time training with self-supervision for generalization under distribution shifts},
  author={Sun, Yu and Wang, Xiaolong and Liu, Zhuang and Miller, John and Efros, Alexei and Hardt, Moritz},
  booktitle={International conference on machine learning},
  pages={9229--9248},
  year={2020},
  organization={PMLR}
}

@article{nado2020evaluating,
  title={Evaluating prediction-time batch normalization for robustness under covariate shift},
  author={Nado, Zachary and Padhy, Shreyas and Sculley, D and D'Amour, Alexander and Lakshminarayanan, Balaji and Snoek, Jasper},
  journal={arXiv preprint arXiv:2006.10963},
  year={2020}
}

@article{schneider2020improving,
  title={Improving robustness against common corruptions by covariate shift adaptation},
  author={Schneider, Steffen and Rusak, Evgenia and Eck, Luisa and Bringmann, Oliver and Brendel, Wieland and Bethge, Matthias},
  journal={Advances in neural information processing systems},
  volume={33},
  pages={11539--11551},
  year={2020}
}

@inproceedings{deng2009imagenet,
  title={Imagenet: A large-scale hierarchical image database},
  author={Deng, Jia and Dong, Wei and Socher, Richard and Li, Li-Jia and Li, Kai and Fei-Fei, Li},
  booktitle={2009 IEEE conference on computer vision and pattern recognition},
  pages={248--255},
  year={2009},
  organization={Ieee}
}

@article{krizhevsky2009learning,
  title={Learning multiple layers of features from tiny images},
  author={Krizhevsky, Alex and Hinton, Geoffrey and others},
  year={2009},
  publisher={Toronto, ON, Canada}
}

@article{sun2019unsupervised,
  title={Unsupervised domain adaptation through self-supervision},
  author={Sun, Yu and Tzeng, Eric and Darrell, Trevor and Efros, Alexei A},
  journal={arXiv preprint arXiv:1909.11825},
  year={2019}
}

@inproceedings{maria2017autodial,
  title={Autodial: Automatic domain alignment layers},
  author={Maria Carlucci, Fabio and Porzi, Lorenzo and Caputo, Barbara and Ricci, Elisa and Rota Bulo, Samuel},
  booktitle={Proceedings of the IEEE international conference on computer vision},
  pages={5067--5075},
  year={2017}
}

@inproceedings{he2016deep,
  title={Deep residual learning for image recognition},
  author={He, Kaiming and Zhang, Xiangyu and Ren, Shaoqing and Sun, Jian},
  booktitle={Proceedings of the IEEE conference on computer vision and pattern recognition},
  pages={770--778},
  year={2016}
}

@article{paszke2019pytorch,
  title={Pytorch: An imperative style, high-performance deep learning library},
  author={Paszke, Adam and Gross, Sam and Massa, Francisco and Lerer, Adam and Bradbury, James and Chanan, Gregory and Killeen, Trevor and Lin, Zeming and Gimelshein, Natalia and Antiga, Luca and others},
  journal={Advances in neural information processing systems},
  volume={32},
  year={2019}
}

@article{ghobrial2023evaluation,
  title={Evaluation Metrics for CNNs Compression},
  author={Ghobrial, Abanoub and Balemans, Dieter and Asgari, Hamid and Reiter, Phil and Eder, Kerstin},
  journal={arXiv preprint arXiv:2305.10616},
  year={2023}
}

@misc{ghobrial2023dira,
      title={DIRA: Dynamic Domain Incremental Regularised Adaptation}, 
      author={Abanoub Ghobrial and Xuan Zheng and Darryl Hond and Hamid Asgari and Kerstin Eder},
      year={2023},
      eprint={2205.00147},
      archivePrefix={arXiv},
      primaryClass={cs.LG}
}

@article{gidaris2018unsupervised,
  title={Unsupervised representation learning by predicting image rotations},
  author={Gidaris, Spyros and Singh, Praveer and Komodakis, Nikos},
  journal={arXiv preprint arXiv:1803.07728},
  year={2018}
}

@inproceedings{gidaris2018dynamic,
  title={Dynamic few-shot visual learning without forgetting},
  author={Gidaris, Spyros and Komodakis, Nikos},
  booktitle={Proceedings of the IEEE conference on computer vision and pattern recognition},
  pages={4367--4375},
  year={2018}
}

@article{Schiappa2023,
author = {Schiappa, Madeline C. and Rawat, Yogesh S. and Shah, Mubarak},
title = {Self-Supervised Learning for Videos: A Survey},
year = {2023},
issue_date = {December 2023},
publisher = {Association for Computing Machinery},
address = {New York, NY, USA},
volume = {55},
number = {13s},
issn = {0360-0300},
url = {https://doi.org/10.1145/3577925},
doi = {10.1145/3577925},
journal = {ACM Comput. Surv.},
month = {jul},
articleno = {288},
numpages = {37}
}

@inproceedings{NEURIPS2022_bcdec1c2,
 author = {Gandelsman, Yossi and Sun, Yu and Chen, Xinlei and Efros, Alexei},
 booktitle = {Advances in Neural Information Processing Systems},
 editor = {S. Koyejo and S. Mohamed and A. Agarwal and D. Belgrave and K. Cho and A. Oh},
 pages = {29374--29385},
 publisher = {Curran Associates, Inc.},
 title = {Test-Time Training with Masked Autoencoders},
 url = {https://proceedings.neurips.cc/paper_files/paper/2022/file/bcdec1c2d60f94a93b6e36f937aa0530-Paper-Conference.pdf},
 volume = {35},
 year = {2022}
}

@article{Song2023,
author = {Song, Yisheng and Wang, Ting and Cai, Puyu and Mondal, Subrota K. and Sahoo, Jyoti Prakash},
title = {A Comprehensive Survey of Few-Shot Learning: Evolution, Applications, Challenges, and Opportunities},
year = {2023},
issue_date = {December 2023},
publisher = {Association for Computing Machinery},
address = {New York, NY, USA},
volume = {55},
number = {13s},
issn = {0360-0300},
url = {https://doi.org/10.1145/3582688},
doi = {10.1145/3582688},
journal = {ACM Comput. Surv.},
month = {jul},
articleno = {271},
numpages = {40}
}

@inproceedings{li2017deeper,
  title={Deeper, broader and artier domain generalization},
  author={Li, Da and Yang, Yongxin and Song, Yi-Zhe and Hospedales, Timothy M},
  booktitle={Proceedings of the IEEE international conference on computer vision},
  pages={5542--5550},
  year={2017}
}

@inproceedings{li2018deep,
  title={Deep domain generalization via conditional invariant adversarial networks},
  author={Li, Ya and Tian, Xinmei and Gong, Mingming and Liu, Yajing and Liu, Tongliang and Zhang, Kun and Tao, Dacheng},
  booktitle={Proceedings of the European conference on computer vision (ECCV)},
  pages={624--639},
  year={2018}
}

@inproceedings{li2019episodic,
  title={Episodic training for domain generalization},
  author={Li, Da and Zhang, Jianshu and Yang, Yongxin and Liu, Cong and Song, Yi-Zhe and Hospedales, Timothy M},
  booktitle={Proceedings of the IEEE/CVF International Conference on Computer Vision},
  pages={1446--1455},
  year={2019}
}

@ARTICLE{Zhou2023,
  author={Zhou, Kaiyang and Liu, Ziwei and Qiao, Yu and Xiang, Tao and Loy, Chen Change},
  journal={IEEE Transactions on Pattern Analysis and Machine Intelligence}, 
  title={Domain Generalization: A Survey}, 
  year={2023},
  volume={45},
  number={4},
  pages={4396-4415},
  doi={10.1109/TPAMI.2022.3195549}}

@article{lopez2017gradient,
  title={Gradient episodic memory for continual learning},
  author={Lopez-Paz, David and Ranzato, Marc'Aurelio},
  journal={Advances in neural information processing systems},
  volume={30},
  year={2017}
}


\end{document}